%% file: main.tex
\documentclass[conference]{IEEEtran}
\IEEEoverridecommandlockouts
\usepackage{cite}
\usepackage{amsmath,amssymb,amsfonts}
\usepackage{algorithmic}
\usepackage{graphicx}
\usepackage{textcomp}
\usepackage{xcolor}
\def\BibTeX{{\rm B\kern-.05em{\sc i\kern-.025em b}\kern-.08em
    T\kern-.1667em\lower.7ex\hbox{E}\kern-.125emX}}
\begin{document}

\title{Automatic Sleep Stage Classification with Cross-modal Self-supervised Features from Deep Brain Signals}

\author{\IEEEauthorblockN{Chen Gong}
\IEEEauthorblockA{\textit{Tsinghua University} \\
gongc16@mails.tsinghua.edu.cn}
\and
\IEEEauthorblockN{Yue Chen}
\IEEEauthorblockA{\textit{Tsinghua University} \\
hellochenyue@foxmail.com}
\and
\IEEEauthorblockN{Yanan Sui}
\IEEEauthorblockA{\textit{Tsinghua University} \\
ysui@tsinghua.edu.cn}
\and
\IEEEauthorblockN{Luming Li}
\IEEEauthorblockA{\textit{Tsinghua University} \\
lilm@tsinghua.edu.cn}
}

\maketitle

\begin{abstract}

The detection of human sleep stages is widely used in the diagnosis and intervention of neurological and psychiatric diseases. Some patients with deep brain stimulator implanted could have their neural activities recorded from the deep brain. Sleep stage classification based on deep brain recording has great potential to provide more precise treatment for patients. The accuracy and generalizability of existing sleep stage classifiers based on local field potentials are still limited. We proposed an applicable cross-modal transfer learning method for sleep stage classification with implanted devices. This end-to-end deep learning model contained cross-modal self-supervised feature representation, self-attention, and classification framework. We tested the model with deep brain recording data from 12 patients with Parkinson's disease. The best total accuracy reached $83.2\%$ for sleep stage classification. Results showed speech self-supervised features catch the conversion pattern of sleep stages effectively. We provide a new method on transfer learning from acoustic signals to local field potentials. This method supports an effective solution for the insufficient scale of clinical data. This sleep stage classification model could be adapted to chronic and continuous monitor sleep for Parkinson's patients in daily life, and potentially utilized for more precise treatment in deep brain-machine interfaces, such as closed-loop deep brain stimulation.

\end{abstract}


\section{Introduction}
Deep brain stimulation (DBS) is a standard surgical treatment for Parkinson's disease (PD) \cite{perlmutter2006deep}. Sleep disorder is a common symptom in PD patients. The specific role of subthalamic nucleus (STN) DBS in sleep was questioned when using continuous high-frequency stimulation. Typical sleep disorders unimproved or even deteriorated after DBS treatment \cite{kim2015rapid}. In recent years, deep brain–machine interfaces have shown the potential for effective and precise treatment systems for the treatment of Parkinson's disease \cite{sui2022deep}. Deep brain-machine interfaces aim to sense deep brain neural activities, and perform targeted interventions on subjects according to different physiological states. In the treatment of Parkinson's disease, closed-loop DBS or adaptive DBS received more attention\cite{parastarfeizabadi2017advances}. It's necessary to detect the sleep state of patients and select the optimal stimulation parameters for treating sleep disorders in the design strategy of closed-loop DBS.

Sleep stage classification is normally done according to the American Academy of Sleep Medicine manual guidelines \cite{berry2018aasm}. The state of sleep is divided into five stages, the awake stage (Wake), the rapid eye movement (REM), and three sleep stages (N1, N2, and N3). Polysomnography (PSG) is a monitoring tool frequently used in the clinic to identify sleep stages \cite{rundo2019polysomnography}. Overmuch wearable sensors of this system negatively affected the sleep quality of the subjects \cite{byun2019first}.

Recorded by contacts in the DBS leads, the local field potential (LFP) is the sum of synchronized activity in a group of neurons and reflects the deep brain activity \cite{brown2005basal}. This deep brain recording system has better integration capabilities without additional external or implanted sensing elements, monitoring sleep stages during daily life without compromising sleep quality by wireless communication.

Studies demonstrated the feasibility of detecting sleep staging based on LFP signals. Christensen et al. acquired LFP signals during sleep in 9 patients with DBS implanted through the percutaneous cable connected to the DBS lead \cite{christensen2019inferring}. They trained an artificial neural network (ANN) to detect three sleep stages (wake, NREM, or REM) based on 30-seconds signals. Chen et al. followed up on 12 patients with Parkinson's disease one month after DBS surgery \cite{chen2019automatic}. This work used entirely in vivo implanted DBS to record LFP wirelessly, and then applied machine learning algorithms (support vector machine, random forest) to distinguish four sleep stages (Wake, N1, N2/N3, REM). All classifiers were based on the LFP frequency domain feature. However, the regularities of frequency-domain energy in different sleep stages were inconsistent in these studies \cite{baumgartner2021basal}. This inconsistency contributes to the lack of robustness in scaling up to new data. 

Feature representation by deep learning gradually became mainstream. The key to deep learning methods is reconciling the contradiction between the need for large-scale data in the training process and the insufficient size of the clinical dataset. Self-supervised learning leverages extensive unlabeled data to learn high-level representations by using the information within or between data. Pretext tasks, such as deletion or modification, force the model to mine intrinsic properties in the data.

This study aimed to design a precise cross-modal sleep stage classification model based on LFP signals. This model contained feature representation, self-attention, and classification framework. We introduced cross-modal self-supervised learning models in the speech domain to train large-scale neural networks on the limited-scale LFP signals dataset. This idea is based on the assumption that speech and LFP signals are highly similar in signal form and time-frequency domain feature representation in specific classification tasks.

We followed up with 12 patients with Parkinson's disease one month after surgery, and recorded the LFP signals the whole night. The end-to-end sleep stage classification model was trained and validated. Compared with classical machine learning methods, our model achieved reliable improvements in accuracy. Results showed the cross-modal self-supervised features improved classification accuracy. In this study, we proposed an interesting perspective on human brain signals. We transferred sleep stage classification of brain signals into auditory perception problems by translating brain signals into speech signal form. This perspective supports cross-modal transfer learning to solve the insufficient scale of clinical data. Our model could be applied to daily sleep monitoring for patients with Parkinson's disease implanted with DBS, and potentially utilized for accurate night-time regulation strategy for closed-loop DBS.

\section{Methods}
\label{sec:methods}

\begin{figure*}[htb]
\centering
 \includegraphics[scale=0.3]{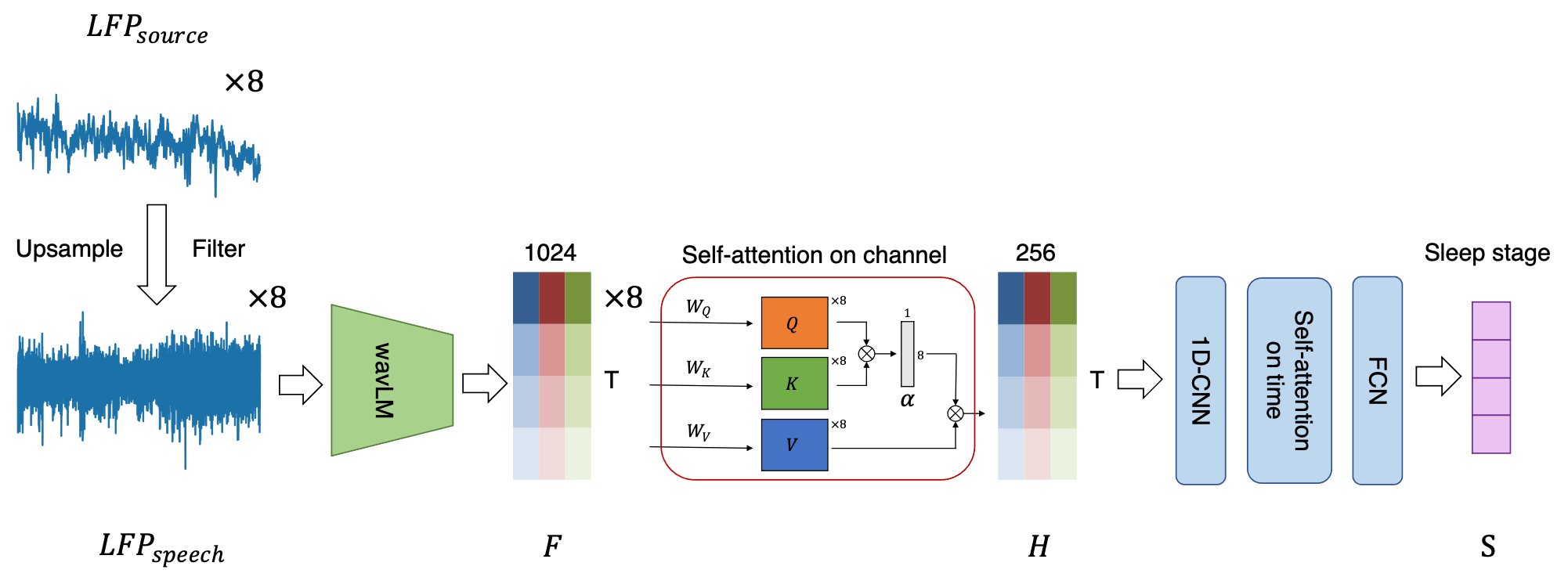}
\caption{Illustration of the cross-modal sleep stage classification model using local field potential signals}
\label{fig:net}
\end{figure*}

In this study, We designed a cross-modal end-to-end model to explore an accurate sleep stage classification method. Fig.~\ref{fig:net} shows the structure of the deep network, composed of feature representation, self-attention, and classification framework. The sleep staging model based on support vector machine proposed by Chen was taken as a benchmark method \cite{chen2019automatic}. We compared the deep end-to-end model with this benchmark model on the same LFP dataset.

\subsection{Dataset}

\begin{figure}[h]
\centering
\includegraphics[scale=0.50]{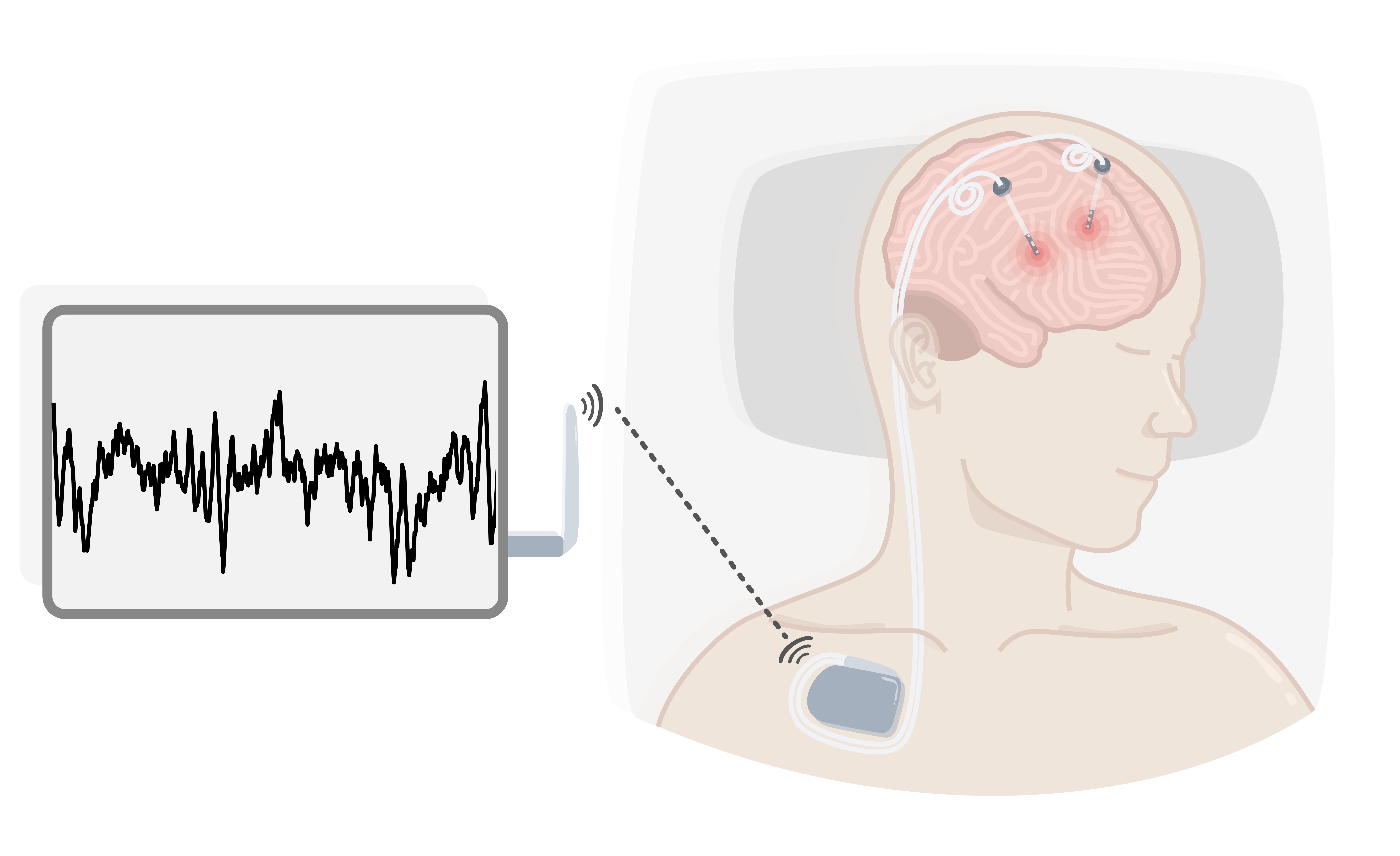}
\caption{Illustration of the local field potential recording under sleep monitoring}
\label{fig:LFP}
\end{figure}

The experiment included 12 patients with Parkinson's disease. All subjects provided written informed consent, and the surgical hospital ethics committee approved the experimental procedures. All subjects were diagnosed with Parkinson's disease and were implanted with bilateral DBS (G106R, PINS, China). This group of patients included seven males aged 40-67 years ($53.29\pm8.99$ years) and five females aged 47-61 years ($57.0\pm5.33$ years). One month after implant surgery, as shown in Fig.~\ref{fig:LFP}, this deep brain signal recording system acquired LFP signals during overnight sleep. To minimize medication effects, subjects discontinued dopaminergic medication 24 hours before the experiment. All LFP signals were sampled at 500 Hz, and eight channels between contact pairs in the electrodes were recorded. In the experiments, subjects' sleep stages were monitored using PSG. According to the American Academy of Sleep Medicine manual guidelines, sleep stages were judged every 30 seconds.

\subsection{Preprocessing}

The LFP signals were split into 5-second segments. To feed the LFP data into the cross-modal self-supervised feature representation framework, we processed the LFP data into a form similar to the speech signal. We used a 0.5Hz high-pass filter to remove motion or system artifacts. Then we took the sample rate as 16000Hz, equivalent to speeding up 32 times. 

\subsection{Feature Representation Framework}

We adopt the wavLM large model to extract speech self-supervised features from LFP signals \cite{chen2022wavlm}. The wavLM model contains a 7-layer convolutional neural network (CNN) encoder and a 24-layer Transformer encoder. The large model was pre-trained with 94,000 hours of large-scale unsupervised English speech. Large-scale unsupervised data from different domains ensured that the wavLM model expressed effective and robust features. Considering the training cost, we froze the parameters of the wavLM model. We took the latent output of each Transformer decoder layer of the model as a feature. We finally acquired 25 feature tensors (wavLM 0-24) for each LFP segment.

\subsection{Self-attention Framework}

The LFP signal is an 8-channel time-series signal, and the channel number indicates the relative position information between the electrode and the target area of brain. Due to the variability brought about by the surgical operation, the distance between the electrode contact and the implantation target was different for each subject. Considering the location information of the channel, We designed the self-attention network of the channel. For the input feature tensor $F=(F_i),i=1,2,\cdots,8$, the query vector, key vector, and value vector were obtained by linear transformation:

$$Q_i=W_Q F_i,K_i=W_K F_i,V_i=W_V F_i$$

The weight score for each channel was: $\alpha _i=softmax(\frac{Q_i\cdot K_i}{\sqrt{d}})$, where $d$ was the dimension of the vector. The output of the channel self-attention framework was:

$$H=\sum_{i=1}^{8} \alpha _i \cdot V_i$$

After training the model on the sleep stage classification task, the channel weight scores reflected the relevance of each channel to the sleep stage. The advantage of the self-attention mechanism is that the trained model automatically selected the best channel for the LFP data of all subjects, which improved the generalization performance in new subjects.

\subsection{Classification Framework} 

We designed a deep classification framework based on deep convolutional networks to gather the temporal characteristics of the output of the channel self-attention network. The model consisted of a 3-layer 1-D CNN with a kernel size of 5. The CNN integrated local information on the input data along the time dimension. Finally, the output of CNN completed the integration of the entire signal through linear mapping and summation, and the model is expressed as:

$$O=CNN(H)$$
$$w_T=softmax(W^{'}_{T} \cdot o_{T})$$
$$S=\sum\limits_{T} w_T \cdot o_T$$

\subsection{Sleep Staging Tasks}

\begin{figure*}[htb]
\centering
\includegraphics[scale=0.3]{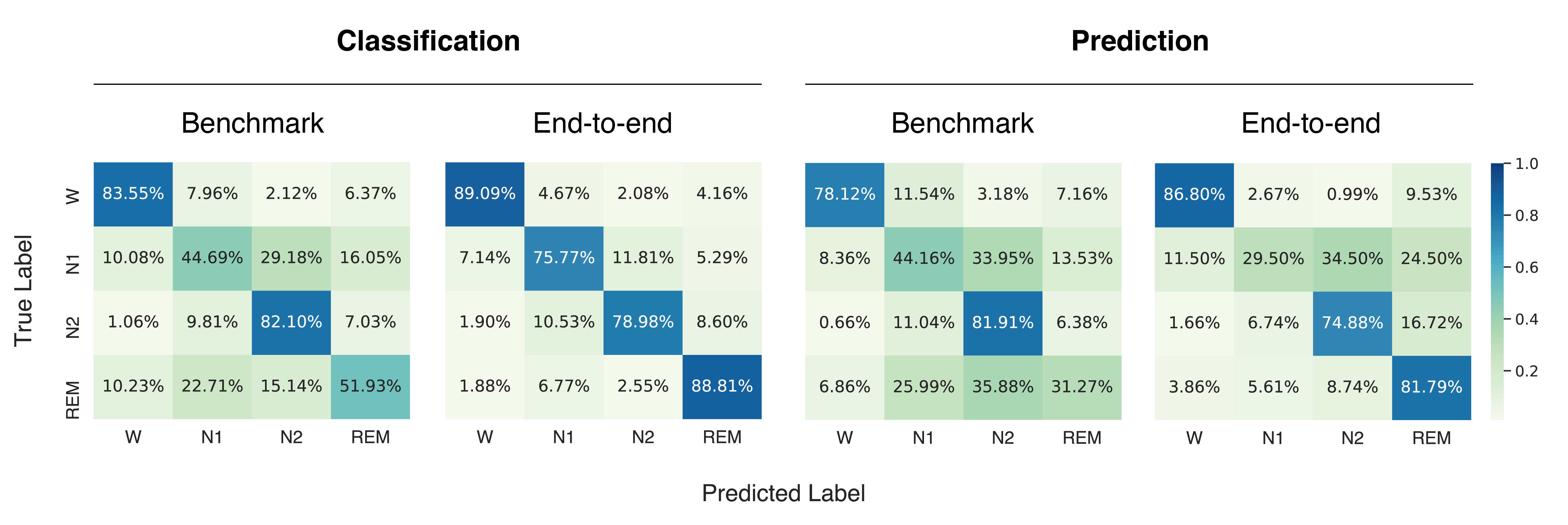}
\caption{Confusion matrices of benchmark and cross-modal end-to-end model}
\label{fig:CM1}
\end{figure*}

We designed two tasks on the LFP dataset. In the classification task, we randomly selected $90\%$ from the data of each label as the training set, and the remaining $10\%$ was the test set. In the prediction task, we selected data segments as the training and test sets in chronological order to simulate more realistic online scenarios. Specifically, we selected the first $90\%$ of time from the data for each label as the training set, and the remaining $10\%$ as the test set. We trained the model to recognize four sleep stages (N2 and N3 were merged).

The data of different sleep stages was unbalanced ($Wake:N1:N2/N3:REM\approx1.5:1:5.5:2.5$), negatively affecting the model's training and validation. We constrained the weight of each class in the training process by using the focal loss function \cite{lin2017focal}:

$$FL(p_t)=-\alpha _t (1-p_t)^{\gamma} log(p_t)$$

\section{Results \& Discussion}

\begin{figure}[h]
\centering
\includegraphics[scale=0.225]{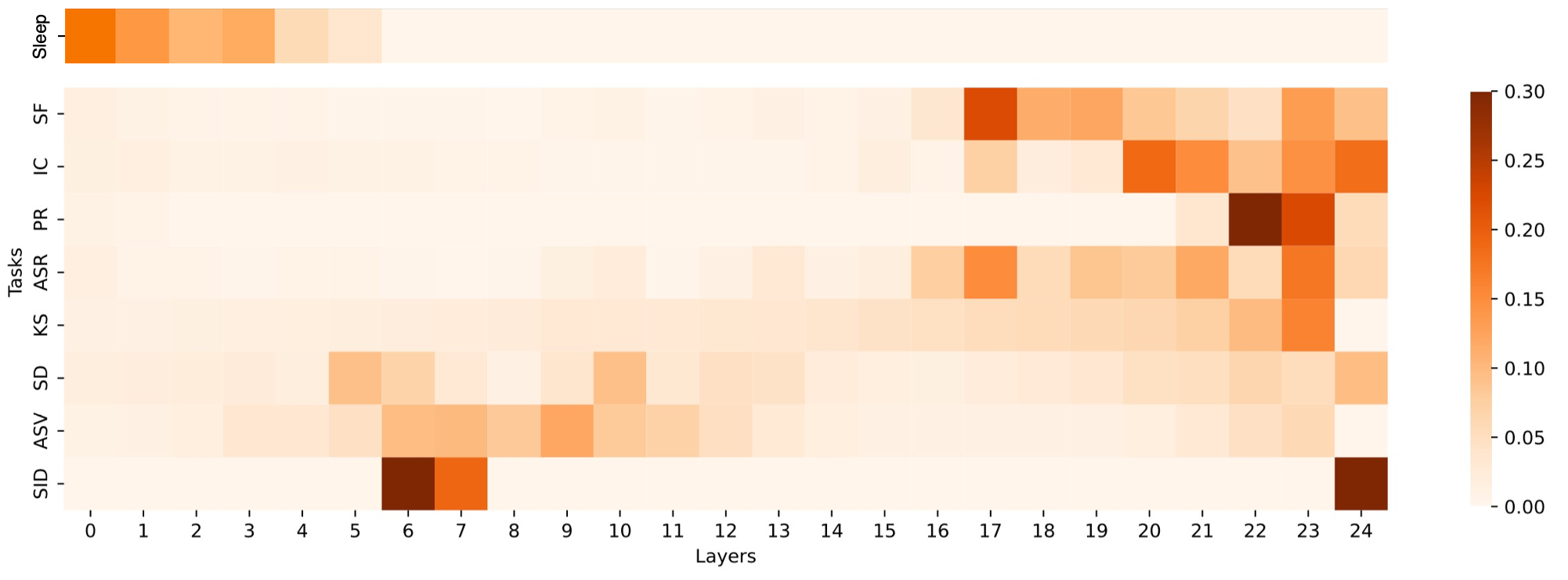}
\caption{Weight analysis of wavLM model on sleep and speech \cite{chen2022wavlm} tasks}
\label{fig:LayW}
\end{figure}

Fig.~\ref{fig:CM1} shows the confusion matrix for benchmark and cross-modal end-to-end model. Compared with the benchmark method, the end-to-end model significantly improved the performance on both sleep staging tasks. Total accuracy reached $83.2\%$ and $68.2\%$ for classification and prediction tasks. The performance was highest for Wake ($89.1\%$ and $86.8\%$), and followed by REM ($88.8\%$ and $81.8\%$), N2 ($79.0\%$ and $74.9\%$), and N1 ($75.8\%$ and $29.5\%$). The deep model tended to predict N1 data as other sleep stages because N1 is the transition between the other three sleep stages in the sleep cycle.

\begin{figure}[h]
\centering
\includegraphics[scale=0.25]{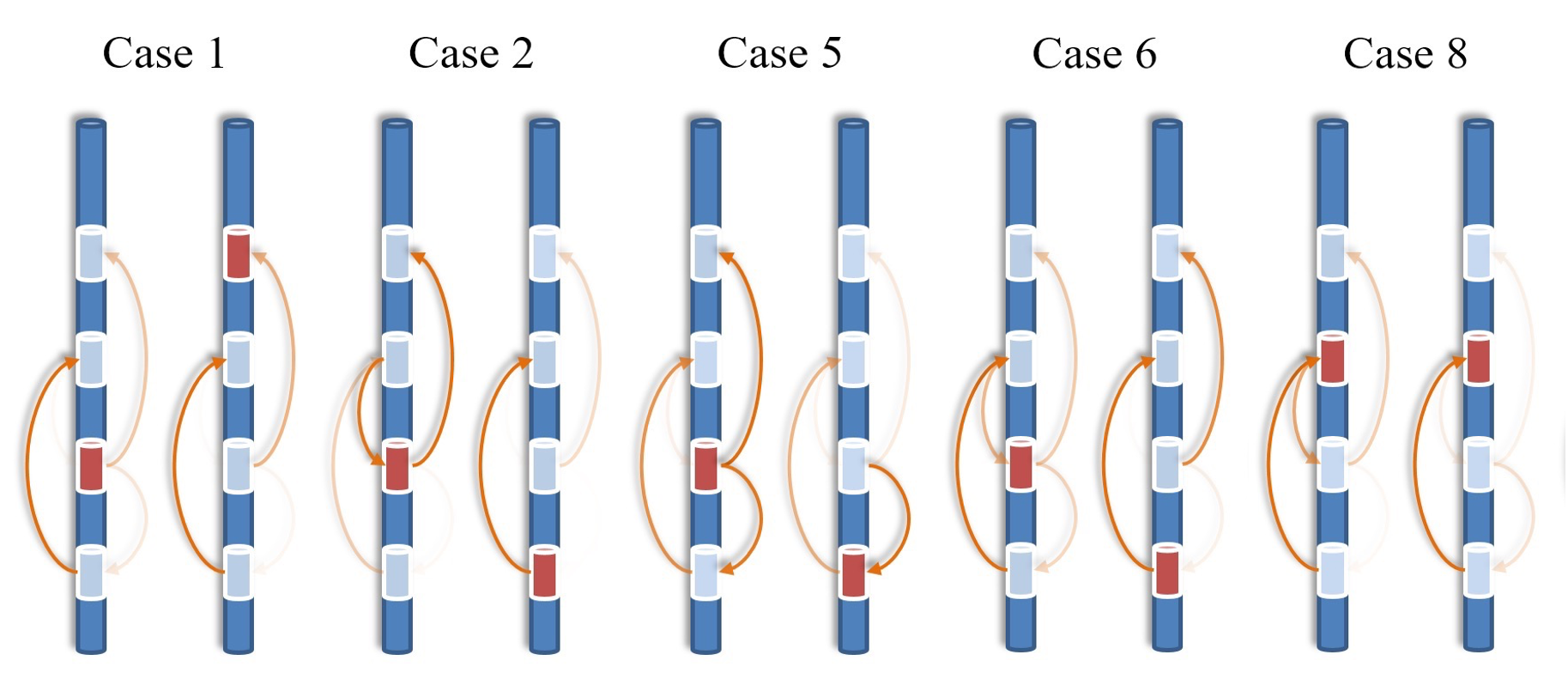}
\caption{Visualizations of therapy electrode and channels with attention weight}
\label{fig:atta_cha}
\end{figure}

We designed ablation experiments to verify the effect of the cross-modal feature representation framework. The frequency domain features of 0-50Hz (same as the benchmark) are combined with self-attention and classification framework for sleep staging. Table.~\ref{tab:acc} showed the total accuracy of three models. The results showed that both the cross-modal feature representation framework and the other two frameworks improved the performance of sleep staging.

\begin{table}[htbp]
\caption{Accuracy of Sleep Staging Models}
\begin{center}
\begin{tabular}{|c|c|c|c|}
\hline
 & \textbf{Benchmark} & \textbf{CNN+self-attention} & \textbf{End-to-End}\\
\hline
\textbf{Classification} &  $65.6\%$ & $70.7\%$ & $83.2\%$ \\
\textbf{Prediction} &  $58.9\%$ & $63.4\%$ & $68.2\%$ \\
\hline
\end{tabular}
\label{tab:acc}
\end{center}
\end{table}

We analyzed the weights of different layers of the wavLM model on the sleep stages classification task. The layer with higher weight significantly contributed to sleep stage classification. Fig.~\ref{fig:LayW} presents the weight of each wavLM layer on sleep and speech \cite{chen2022wavlm} tasks. Unlike the downstream speech task, the model had significantly higher weights in the shallow layers. We inferred physiologically relevant low-level features in LFP signals were critical factors in distinguishing different sleep stages.

In the process of training the deep model, the data drove the self-attention sub-network to characterize the weights of channels of the LFP signal. The weights showed a consistent distribution across subjects. We averaged the weights of the eight channels per subject and compared them with the treatment electrodes. As shown in Fig.~\ref{fig:atta_cha}, the red electrodes represented therapy electrodes; the red connecting lines represented the bipolar of the acquisition channel. The darker the color, the higher the weight score. In most subjects, the high-weighted acquisition channel and therapy electrodes exhibited two patterns: the therapy electrode was one of the bipolar of the acquisition channels; the bipolar of the acquisition channel was the adjacent electrode of the therapy electrode. We inferred there is a functional connection between sleep and treatment in PD patients.

\section{Conclusions and Future Work}
\label{sec:floats}

In this study, we built a cross-modal end-to-end model of sleep stage classification based on local field potential signals. The performance of this deep model was significantly improved over classical machine learning methods. Unlike previous studies on human brain signals, our sleep stage classification model identified brain signals via "listening". We observed the similarity between LFP and speech signals, introducing a speech self-supervised model to extract deep features of LFP. This work extended the boundaries of cross-modal transfer learning in temporal physiological signals. 

Our method provides a fully implanted sleep stage detection strategy for PD patients, which can be applied in closed-loop deep brain–machine interfaces treatment systems. Considering the variability of LFP signals after long-term implant surgery, we will build a longitudinal LFP sleep dataset and validate the deep learning model. A closed-loop DBS system for Parkinson's patients based on LFP sleep staging will be designed and verified. After collecting the LFP signal, the device produces sleep staging results. Then the processor makes treatment parameter adjustment decisions. A precise sleep stage classification model would help us develop more effective treatments for Parkinson's disease.

\section{Acknowledgment}

This work was financially supported by STI 2030-Major Projects (2022ZD0209400) and LG-QS-202202-10. We thank Jingying Ye, Guoping Yin, Xing Cao and Yuhuan Zhang for sleep experiment; Jianguo Zhang, Yi Guo and Shujun Xu for DBS surgery; Hongwei Hao for experimental design. We thank Hongjiang Liu for the experimental schematic.

\input{main.bbl}

\bibliographystyle{IEEEtran}

\end{document}